\documentclass[journal,twoside,web,hidelinks]{ieeecolor}
\usepackage{caption}
\usepackage{subcaption}
\usepackage{generic}
\usepackage{lipsum}
\usepackage{amsmath,amssymb,amsfonts}
\usepackage{graphicx}
\usepackage[disable]{todonotes} % TODO: add disable to remove todo notes
\usepackage{textcomp}
\usepackage{orcidlink}
\usepackage{multirow}
\usepackage{algorithm}
\usepackage{algpseudocode}
\usepackage[natbib=true,style=ieee,sorting=none,mincitenames=1,maxcitenames=2]{biblatex}
\def\BibTeX{{\rm B\kern-.05em{\sc i\kern-.025em b}\kern-.08em
    T\kern-.1667em\lower.7ex\hbox{E}\kern-.125emX}}
\AtBeginBibliography{%
    \DeclareFieldFormat{labelnumberwidth}{\ifcategory{new_citations}%
        {\color{red}\mkbibbrackets{#1}}%
        {\mkbibbrackets{#1}}%
    }}
\AtEveryBibitem{\ifcategory{new_citations}
    {\color{red}}
    {}}

\begin{document}
\title{Zero-Shot Medical Phrase Grounding with Off-the-shelf Diffusion Models}
\author{Konstantinos Vilouras \orcidlink{0009-0003-5910-9748}, Pedro Sanchez \orcidlink{0000-0003-2435-3049}, Alison Q. O'Neil \orcidlink{0000-0001-8371-0603}, and Sotirios A. Tsaftaris \orcidlink{0000-0002-8795-9294}, \IEEEmembership{Senior Member, IEEE}
\thanks{Manuscript submitted for review on 6 November 2024. This work was supported by the University of Edinburgh, the Royal Academy
of Engineering and Canon Medical Research Europe by PhD studentships to
Konstantinos Vilouras and Pedro Sanchez. S.A. Tsaftaris also acknowledges the support of Canon
Medical and the Royal Academy of Engineering and the Research Chairs and
Senior Research Fellowships scheme (grant RCSRF1819\textbackslash 8\textbackslash 25), and the UK’s Engineering
and Physical Sciences Research Council (EPSRC) support via grant
EP/X017680/1. We also acknowledge support of the UKRI AI programme, and the Engineering and Physical Sciences Research Council, for CHAI - Causality in Healthcare AI Hub [grant number EP/Y028856/1].}
\thanks{Konstantinos Vilouras, Pedro Sanchez, Sotirios A. Tsaftaris are with the School of Engineering, University of Edinburgh, Edinburgh EH9 3FB, United
Kingdom (e-mail: \href{mailto:konstantinos.vilouras@ed.ac.uk}{konstantinos.vilouras@ed.ac.uk}, \href{mailto:pedro.sanchez@ed.ac.uk}{pedro.sanchez@ed.ac.uk}, \href{mailto:s.tsaftaris@ed.ac.uk}{s.tsaftaris@ed.ac.uk}). }
\thanks{Alison O'Neil is with Canon Medical Research Europe Ltd., Edinburgh EH6 5NP, United Kingdom (e-mail: \href{mailto:alison.oneil@mre.medical.canon} {alison.oneil@mre.medical.canon}) and with the School of Engineering, University of Edinburgh, Edinburgh EH9 3FB, United
Kingdom.}
}

\maketitle

\begin{abstract}
Localizing the exact pathological regions in a given medical scan is an important imaging problem that traditionally requires a large amount of bounding box ground truth annotations to be accurately solved. However, there exist alternative, potentially weaker, forms of supervision, such as accompanying free-text reports, which are readily available. 
%and also arguably provide more detailed information. 
The task of performing localization with textual guidance is commonly referred to as phrase grounding. In this work, we use a publicly available Foundation Model, namely the Latent Diffusion Model, to perform this challenging task. This choice is supported by the fact that the Latent Diffusion Model, despite being generative in nature, contains cross-attention mechanisms that implicitly align visual and textual features, thus leading to intermediate representations that are suitable for the task at hand. In addition, we aim to perform this task in a zero-shot manner, i.e., without any training on the target task, meaning that the model's weights remain frozen. To this end, we devise strategies to select features and also refine them via post-processing without extra learnable parameters. We compare our proposed method with state-of-the-art approaches which explicitly enforce image-text alignment in a joint embedding space via contrastive learning. Results on a popular chest X-ray benchmark indicate that our method is competitive with SOTA on different types of pathology, and even outperforms them on average in terms of two metrics (mean IoU and AUC-ROC). Source code will be released upon acceptance at \url{https://github.com/vios-s}.
\end{abstract}


\section{Introduction}
\label{sec:introduction}
\IEEEPARstart{T}{he} rapid success of deep learning over the last few years has led to powerful data-driven models being deployed in real-world scenarios. Recently, by taking advantage of the scaling properties of popular deep learning methods both in terms of learnable parameters and training data, we witness the era of Foundation Models (FMs) \cite{bommasani2021opportunities}, i.e., large-scale neural networks that were trained on massive amounts of data. FMs have unprecedented capabilities: they can be readily applied to a wide variety of tasks as off-the-shelf solutions, or they can serve as a robust basis for training models for specific, potentially unseen, tasks and, plausibly, modalities (e.g., transferring knowledge from natural images to the medical domain). Among their many benefits, FMs provide machine learning practitioners and researchers with a universal tool that enables the widespread application of data-driven solutions to multiple scientific fields, as well as the development of a sound theoretical framework around well-known deep learning methodologies.

\begin{figure}[!t]
\centerline{\includegraphics[width=\columnwidth]{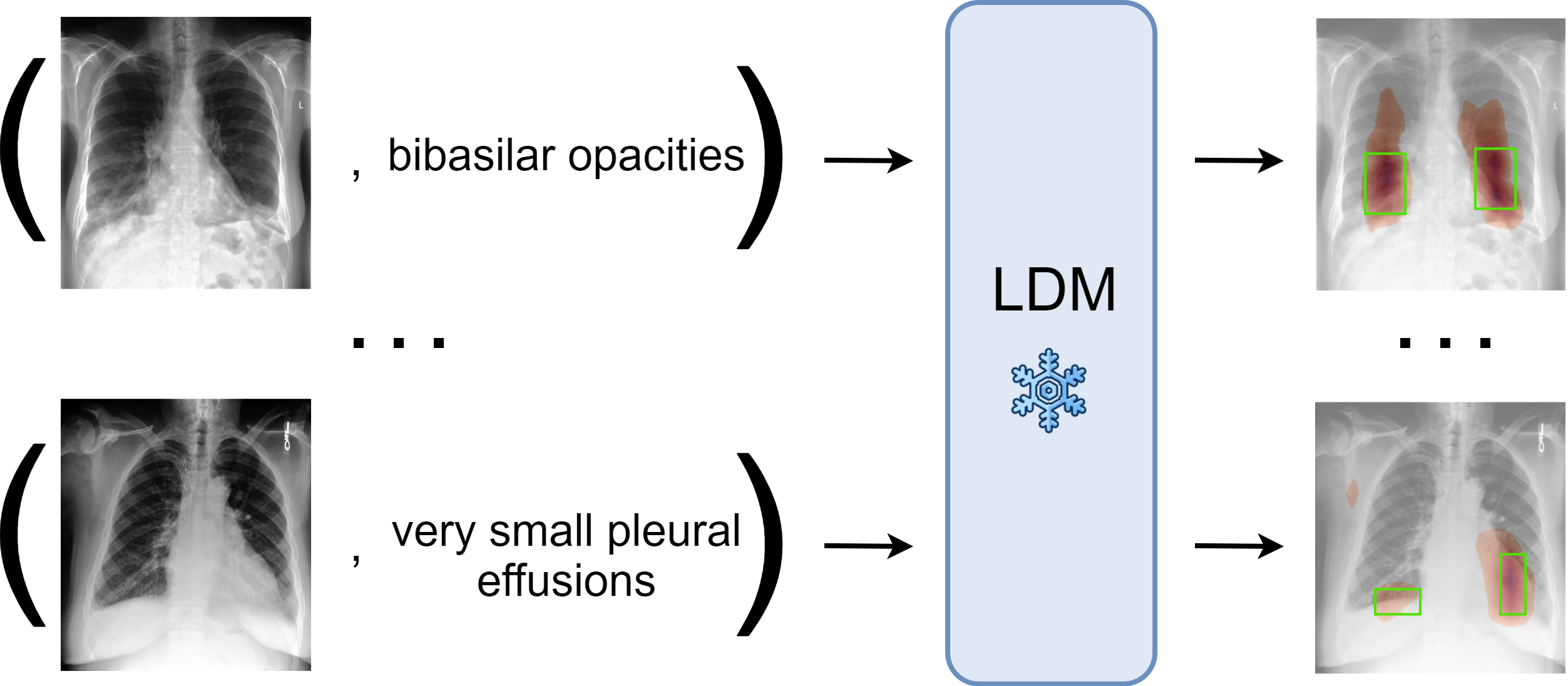}}
\caption{High-level description of the zero-shot phrase grounding task. Given input pairs of an image (chest X-ray) and its accompanying text prompt, we leverage cross-modal feature alignment mechanisms within a frozen Latent Diffusion Model (LDM) to extract heatmaps, which indicate the regions where image and text are maximally aligned. Then, we evaluate the generated heatmaps based on ground truth bounding boxes (shown in green) for pathology detection. Our method, thus, is an illustration of using pre-trained LDMs for downstream applications in a zero-shot setting.}
\label{fig1}
\end{figure}

In this work, we investigate a specific type of FM: the Latent Diffusion Model (LDM) \cite{rombach2022high}. The LDM belongs to the class of probabilistic models and is considered one of the most groundbreaking methods for image synthesis. The versatile design of the LDM has significantly contributed towards its success, as external sources of information (e.g., text, segmentation masks, or any other type of a representation) can be easily incorporated into the model without any architectural changes. Here, we draw inspiration from research studies showing that diffusion models can solve downstream tasks such as classification \cite{li2023your,clark2024text} and segmentation \cite{baranchuk2021label} with little to no additional fine-tuning on target data. We attempt to validate this finding i.e., the re-usability of diffusion models for downstream tasks, in the medical imaging domain, using a model pretrained on medical images (i.e. chest X-rays) with readily available associated text (i.e. radiology reports) as conditioning.

In order to leverage a model trained with text as conditioning, we reframe the objection detection task as one of \emph{phrase grounding}. In the context of diffusion models, instead of training a model that predicts bounding box coordinates \cite{chen2023diffusiondet}, we propose to leverage cross-modal feature fusion mechanisms within an off-the-shelf LDM to directly perform phrase grounding, relying on the model having learned accurate fine-grained image-text alignment.
%  Moreover, since the LDM was pre-trained on image generation, we operate in a zero-shot task transfer setting.
Moreover, we consider the case of \textit{zero-shot task transfer} \cite{pal2019zero} in our work, since the LDM's pre-training task (text-guided image generation) differs from the target task (phrase grounding). However, in contrast to \cite{pal2019zero} which requires further training using supervision across multiple tasks, we use an off-the-shelf model (LDM) with frozen weights to solve the task at hand.

Prior works have defined the phrase grounding task in multiple ways. Along with input image-text pairs, most methods further require either ground truth bounding boxes \cite{zhao2022word2pix} or object detection models \cite{liu2024vgdiffzero} during training, allowing test-time selection of the region proposal that most closely matches the input text. On the contrary, we recognise that manual ground truth bounding box annotations and external pathology detection models are typically difficult to acquire. Instead, we opt for an end-to-end system that extracts relevant information from natural language (e.g., location and severity modifiers of the underlying pathology) and, in turn, associates clinical findings with visual features corresponding to specific image regions. A high-level overview of our method for the phrase grounding task is shown in Fig.~\ref{fig1}.

There are emergent properties of the LDM that are useful for the given task. First, the core of the LDM is a U-Net architecture \cite{ronneberger2015u} which, in turn, is equipped with inductive biases (e.g., multi-scale hierarchical feature learning) suitable for localisation tasks such as phrase grounding. This differentiates our method from other baselines that use image classification models such as ResNets \cite{boecking2022making,bannur2023learning}. In addition, the LDM offers a sophisticated feature fusion mechanism: fusion layers (cross-attention) are incorporated at multiple levels of the architecture, and visual features evolve over time via the diffusion process. Therefore, this mechanism is expected to yield more refined representations compared to late fusion alone \cite{boecking2022making}.
% established strategies in multi-modal learning such as 

We perform extensive experiments on an established phrase grounding benchmark dataset, i.e., MS-CXR \cite{boecking2022making}. The results suggest that, despite being inherently a generative model, the LDM has learned high quality features for the task at hand. Our proposed approach, which tackles the extreme case where no additional fine-tuning is performed (we refer to it as a \textit{zero-shot} scenario), yields, perhaps surprisingly, a highly competitive method that proves to be state-of-the-art in terms of two metrics (mean IoU and AUC-ROC) on average across 8 pathology labels.

Overall, our \textbf{contributions} can be summarised as follows:
\begin{itemize}
    \item Building on the Latent Diffusion Model architecture, we gather semantically meaningful visual features both from multiple timesteps of the diffusion process and from various cross-attention layers. We target those layers since they inherently align information from the visual and textual stream, thus being suitable for the phrase grounding task.
    \item In contrast to conventional sampling methods for Latent Diffusion Models that utilise classifier-free guidance, our approach involves sampling from the unconditional model, while the conditional model (conditioned on a text input) is used merely to extract cross-attention maps.
    %text information is incorporated in a separate path merely to extract cross-attention maps.
    \item We perform extensive experiments on a medical dataset that provides ground truth bounding boxes for evaluation and test our method against several strong baselines. Results indicate that our method shows competitive performance against, and even exceeds in two metrics, state-of-the-art methods without any fine-tuning strategies.
    \item We perform an ablation study to justify the hyperparameter choices in our system.
    \item We further qualitatively analyze  our proposed method, as well as the strongest available baselines, to provide useful insights for those approaches. 
\end{itemize}

\section{Related Work}
\label{sec:relatedwork}
We now summarise the most relevant research. We start by mentioning prior influential works showing that diffusion models can be effectively applied to multiple downstream tasks in a zero- or few-shot setting. Next, due to their similarity to our approach, we also present methods tailored for image editing tasks that utilise Latent Diffusion Models. Then, we briefly discuss some of the most popular methods for phrase grounding in natural RGB images. Finally, we shift our focus to approaches related to the medical imaging domain, against which we compare our proposed method. 

\subsection{Downstream Application of Diffusion Models}
Diffusion models have been successfully applied to a wide range of tasks. For instance, in the context of medical imaging, there exist works that train diffusion models from scratch to perform lesion localisation \cite{wolleb2022diffusion,sanchez2022healthy}, anomaly detection \cite{pinaya2022fast}, and counterfactual generation \cite{bedel2023dreamr,sanchez2022diffusion}. Recent studies have also revealed the ability of diffusion models to perform fairly well on downstream tasks with minimal supervision. For example, for classification tasks it has been shown that the posterior $p(\mathbf{c}|\mathbf{x})$ for all candidate classes $\mathbf{c}$ can be estimated from a diffusion model's residual errors at a given timestep without any requirement for further hyperparameters or training \cite{li2023your, clark2024text}. Similarly, for segmentation, \citet{baranchuk2021label} use intermediate visual representations extracted from a diffusion model. In this case, a few additional labelled images are required to train a shallow network that outputs pixel-wise predictions. In another work, \citet{zhao2023unleashing} experiment with the text feature extraction pipeline, as well as the choice of intermediate visual features, and train lightweight task-specific models for segmentation and depth estimation, respectively.

\subsection{Manipulating Attention in Latent Diffusion Models}

Closely related to our approach, yet developed for the task of image editing, recent studies use attention maps extracted from a Latent Diffusion Model to control the generated images. For example, \citet{hertz2022prompt} showed that it is possible to apply global and local edits to an image by processing the cross-attention maps generated via source and target prompts, respectively. \citet{patashnik2023localizing} aim to vary the shape of a target object in a generated image, and they use both cross- and self-attention maps to better preserve the shape and appearance for the rest of the image. Conversely, \citet{tumanyan2023plug} argue that only convolutional and self-attention features are useful for editing, since localised visual information is not described in text prompts.

\subsection{Visual Phrase Grounding}
Phrase grounding is a cross-modal reasoning task referring to the spatial localisation of objects present in an image given a relevant text description. Due to the widespread availability of image-text data sources, there already exist large-scale end-to-end models for the natural image domain.

In a discriminative learning scenario, ViLBERT \cite{lu2019vilbert} introduces the concept of co-attention, i.e., exchanging information between modalities within the transformer layers, and these co-attention mechanisms can be used to directly relate visual and textual tokens. Early fusion strategies are also adopted by MDETR \cite{kamath2021mdetr} which enforces image-text alignment with appropriate learning objectives, and GLIP \cite{li2022grounded} in which image and language encoders are simultaneously trained to correctly assign a word to a specific image region. 

In a generative learning scenario, \citet{chen2023language} train a text-guided diffusion model to gradually recover ground truth bounding boxes from their noise-perturbed versions.
On the other hand, \citet{tang2022daam} focus on a standard image generation model and demonstrate how various linguistic aspects of the input text prompt affect the generated image, by extracting and visualising the cross-attention heatmaps; this allows interpretation of stable diffusion's image generation process. They further evaluate use of the heatmaps for segmenting the objects, a localisation task which has parallels with phrase grounding, albeit the target image is a result of the text prompt rather than a parallel input. This approach is similar in principle to ours, but for the opposite process of denoising (corresponding to image generation). \

\subsection{Medical Visual Phrase Grounding}
In the medical imaging domain, phrase grounding is considered a difficult task due to the inherent variation in textual information; radiologists commonly use domain-specific terms, describe the absence of pathological findings (e.g. ``\emph{No pneumothorax or pleural effusion}''), or use phrases that convey a level of uncertainty (e.g. ``\emph{Blunting of the right costophrenic angle is consistent with a small right pleural effusion}''.

Earlier works have taken different approaches. \citet{bhalodia2021improving} extract pneumonia-related attributes from radiology reports, while a pre-trained bounding box detector is used to extract regions of interest (ROIs) and their associated features. Then, using both streams of information, their system is trained to correctly classify attributes from visual features, as well as to maximise the similarity for a given image-text pair.

End-to-end discriminative methods to date are largely fully supervised \cite{chen2023medical,xu2023learning} or rely on self-supervised contrastive formulations \cite{dawidowicz2023limitr,boecking2022making,bannur2023learning}. Specifically, \citet{chen2023medical} train a vision-language Transformer model to directly predict bounding boxes, whereas \citet{xu2023learning} gather publicly available labelled chest X-ray datasets and train a single model via multi-task learning. LIMITR \cite{dawidowicz2023limitr} is a self-supervised method that aligns local cross-modal representations that are further weighted via learnable significance scores. Lastly, BioViL \cite{boecking2022making} is an end-to-end model with a BERT text encoder finetuned on radiology reports that is optimised via both global and local cross-modal contrastive losses. \citet{bannur2023learning} extend the BioViL model to support longitudinal information across patients, and the resulting system (BioViL-T) achieves state-of-the-art performance on medical phrase grounding.

In contrast, we adopt a generative approach to this task. A concurrent work also applies diffusion models to phrase grounding \cite{dombrowski2023tradeoffs}. Our approach differs in several ways: First, we use a pre-trained, publicly available diffusion model and propose mechanisms to perform phrase grounding whilst keeping the model frozen, whereas \cite{dombrowski2023tradeoffs} focuses on training the model from scratch. Second, to define these mechanisms, we pay attention to how we select cross-attention layers and timesteps, whereas \cite{dombrowski2023tradeoffs} simply average across both time and layers. Finally, 
%more on the model's training part and merely reports phrase grounding results, while we provide strategies to achieve highly competitive performance for the task at hand. Moreover, 
we compare our method with more recent state-of-the-art and other, recently proposed, baselines. Nevertheless, the findings of \cite{dombrowski2023tradeoffs} are beneficial for our work. They show that learning both text and visual encoders simultaneously severely degrades performance. In addition, regarding the textual information that is available for phrase grounding, they show that using sentences from the original radiology reports (as is the case, for example, for the MS-CXR \cite{boecking2022making} dataset) leads to the best overall results, outperforming both synthetically generated text (with ChatGPT) and the simplest case of using the class string as the input prompt.

\section{Methodology}
\label{sec:methodology}

\begin{figure*}[!t]
\centering
\includegraphics[width=0.95\textwidth,keepaspectratio]{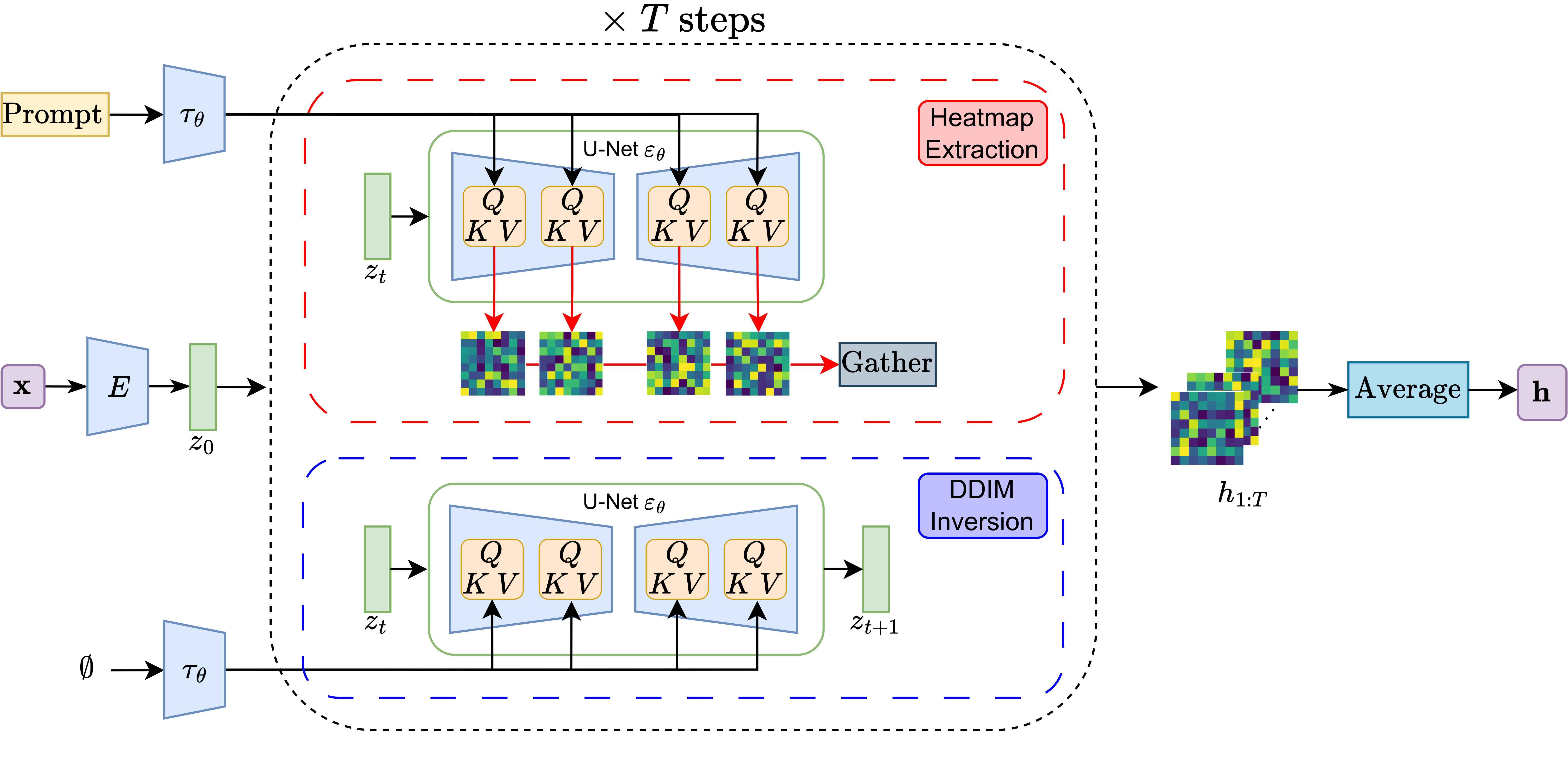}
\caption{Overview of our proposed phrase grounding pipeline based on the Latent Diffusion Model \cite{rombach2022high}. The input image-text pair is first processed via the encoders $E$ and $\boldsymbol{\tau_\theta}$, respectively. Then, at each timestep of the diffusion process $t=1,...,T$, we gather cross-attention maps from the U-Net $\boldsymbol{\epsilon_\theta}$. The output heatmap $\mathbf{h}$ is generated by averaging the gathered attention maps.
}
\label{fig2}
\end{figure*}

We adopt the following notation: x is a scalar, $x$ denotes a vector, $\mathbf{x}$ is a tensor, $\mathcal{X}$ denotes a vector space, $\mathbf{f}: \mathcal{X}\rightarrow \mathcal{X}$ is the mapping between two vector spaces that is performed by a neural network $\mathbf{f}$, and $\mathbf{f_\theta}$ refers to the total number of learnable parameters (weights) in the network. We also consider a dataset $\mathcal{D} = \{\mathbf{x_i}, p_i\}_{i=1}^N$ with a total of $N$ image-text pairs. For simplicity, we will next present our pipeline while considering a single input $(\mathbf{x}, p)$; the extension to a batch is trivial.

\subsection{Overview of the Latent Diffusion Model}
%The Latent Diffusion Model (LDM) \cite{rombach2022high} belongs to the class of probabilistic models and is considered one of the most groundbreaking methods for image synthesis. The versatile design of the LDM has significantly contributed towards its success, as external sources of information (e.g., text, segmentation masks, or any other type of a representation) can be easily incorporated into the model without any architectural changes. Here,
We follow the standard setup of the LDM architecture as proposed in \cite{rombach2022high}. The core components of the LDM are as follows. First, the visual encoder $\mathbf{E}: \mathcal{X}\rightarrow \mathcal{Z}$ yields a compressed latent representation $z_0$ of the input image $x$. Then, in the first part of the diffusion process (also called the forward process), Gaussian noise is added to $z_0$. This process is iterative, meaning that noise is repeatedly added over time to $z_t$ ($t=1, ..., T$), leading to a final latent representation $z_T$ that is pure noise. Next, in the second part of the diffusion process (i.e., the reverse process), the goal is to repeatedly denoise the latent representations $z_t$ ($t=T, ...,1$) until $z_0$ is recovered. The denoising task is carried out by a U-Net model $\boldsymbol{\epsilon_\theta} : \mathcal{Z}\rightarrow \mathcal{Z}$ that predicts the noise introduced at timestep $t$. Finally, a visual decoder $\mathbf{D}: \mathcal{Z}\rightarrow \mathcal{X}$ generates the reconstructed image $\mathbf{\tilde{x}}$.
%More specifically, via the encoder $\mathbf{E}: \mathcal{X}\rightarrow \mathcal{Z}$, the input image $\mathbf{x}$ is first encoded into a latent feature vector $z$ which, in turn, is progressively corrupted with Gaussian noise via a deterministic forward diffusion process for a total of T timesteps. Then, for each timestep t, a denoising U-Net $\boldsymbol{\epsilon_\theta} : \mathcal{Z}\rightarrow \mathcal{Z}$ is applied to the noisy latent $z_t$ to recover the cleaner estimate at the previous step $z_{t-1}$. After the reverse diffusion process, the recovered global latent $z$ is mapped to the original pixel space via a decoder $\mathbf{D}: \mathcal{Z}\rightarrow \mathcal{X}$, leading to the reconstructed input $\mathbf{\tilde{x}}$. An overview of the LDM pipeline is depicted in Fig.\ \ref{fig2}.

\subsection{Image-Text Alignment via Cross-Attention}
To aid reconstruction, information about the image $\mathbf{x}$ in text form is also injected in the U-Net model $\boldsymbol{\epsilon_\theta}$ via the cross-attention layers, which ultimately capture interactions between the visual and textual streams. Concretely, the prompt $p$ that accompanies the image $\mathbf{x}$ is first split into $S$ tokens and then passed through a text encoder $\boldsymbol{\tau_\theta}$. The resulting text features serve as context in each cross-attention layer of the U-Net, i.e., they are used to generate keys $\mathbf{k}$ and values $\mathbf{v}$, whereas queries $\mathbf{q}$ are extracted from visual features.
%Intuitively, queries $\mathbf{q}$ (resp.\ keys $\mathbf{k}$) provide a high-level description of the visual (resp.\ textual) content per spatial location (resp.\ token), while values $\mathbf{v}$ contain detailed features for each token.
Intuitively, queries $\mathbf{q}$ provide a high-level description of the visual content per spatial location, keys $\mathbf{k}$ provide a high-level description of the textual content per token, and values $\mathbf{v}$ contain more detailed textual features per token.
The cross-attention operation is defined in Eq.\ \eqref{eq:attn} below, where $\mathbf{P_K}$ and $\mathbf{P_V}$ denote linear projection layers for the keys and values, respectively, and $d$ is a scaling factor. 

%for the attention operation as defined below, where $\mathbf{q}$ refers to the queries extracted from the visual features, $\mathbf{P_K}$ and $\mathbf{P_V}$ are projection layers for the keys and values, respectively, and $d$ is a scaling factor,
%
\begin{equation}
\begin{aligned}
  \mathbf{k} = \mathbf{P_K}(\tau_\theta(p)) &\qquad \mathbf{v} = \mathbf{P_V}(\tau_\theta(p)) \\
  \underbrace{A_p = \text{softmax}(\frac{\mathbf{q}\mathbf{k}^\top}{\sqrt{d}})}_\text{cross-attention map} &\qquad \text{Attention}(\mathbf{q},\mathbf{k},\mathbf{v}) = A_p \cdot \mathbf{v}. 
\end{aligned}
\label{eq:attn}
\end{equation}
At each layer and timestep, cross-attention maps $A_p \in \mathbb{R}^{S\times B\times B}$ are computed according to Eq.\ \eqref{eq:attn}, where $S$ is the number of prompt tokens and $B$ is the spatial size (either 16 or 32, depending on the layer's level in the U-Net architecture). Let $\text{N}$ denote the total number of cross-attention layers in the U-Net and $\text{T}$ the number of diffusion timesteps. That leaves us with $N \times T$ attention maps in total, where each one varies depending on the visual features (low-level features or higher-level semantics) and the noise level in each latent $z_t$. The extracted cross-attention maps are then used for phrase grounding.

\subsection{Zero-shot Phrase Grounding with the LDM}
\label{sec:ldmgrounding}
We use a pre-trained LDM to perform zero-shot phrase grounding. More specifically, given an image and a prompt as input, we extract intermediate cross-attention maps from the U-Net and aggregate them into a heatmap $\mathbf{h}$. In the ideal case of perfect image-text alignment, $\mathbf{h}$ will be highly activated on the regions of interest specified by the prompt. An overview of our approach is presented in Fig.\ \ref{fig2}.
% Different timesteps $t$ in the diffusion process lead to different semantic information in the representation of the Unet \cite{baranchuk2021label}. We estimate cross-attention maps for multiple timesteps using DDIM inversion \cite{dhariwal2021diffusion}. Then, we aggregate the attention maps to generate a heatmap. In the ideal case of perfect image-text alignment, $\mathbf{h}$ will be highly activated on the regions of interest specified by the prompt. The red block in Fig.\ \ref{fig1} shows the process for generating the heatmap $\mathbf{h}$. 

Specifically, our method is based on DDIM inversion \cite{dhariwal2021diffusion} which maps input features $z_0$ into noise $z_T$. Note that we avoid classifier-free guidance during sampling as it would increase the accumulated error introduced by DDIM inversion \cite{mokady2023null}. To this end, we sample $z_{t+1}$ from the LDM using only the unconditional inputs: the latent representation $z_{t}$ and a null prompt (the empty string $\emptyset$). However, we gather the cross-attention maps using the previously sampled $z_{t}$ and the conditioning input (prompt $p$). Our method is summarised in Alg.\ \ref{alg:ddim_inversion}.
%  (i.e., the inferred noise $z_T$ does not map back to the input $z_0$ via the forward diffusion process)
%DDIM inversion allows running a diffusion model in the forward direction without adding noise to the image \cite{dhariwal2021diffusion}. Similar to classifier-free guidance, we run the diffusion model on the latent space with the prompt $p$ and with an empty conditioning $\emptyset$. Only the results of the empty conditioning $\epsilon_\emptyset$ are used to run DDIM inversion. Only the cross-attention maps from the latent space with the prompts $A_{p,t}$ are used for phrase grounding. This technique allows measuring the effect of a prompt in the image without intervening on the images \cite{mokady2023null}. See Alg.\ \ref{alg:ddim_inversion} for a description of this procedure with DDIM inversion. Note that }

\begin{algorithm}
\caption{Extracting cross-attention maps via DDIM inversion.}
\label{alg:ddim_inversion}
\begin{algorithmic}[1]
\Require image $x$, prompt $p$, encoder $\mathbf{E}$, U-Net $\boldsymbol{\epsilon_\theta}$, timesteps $T$, noise schedule parameters $\alpha_t$
\State $z_0 \leftarrow \mathbf{E}(x)$ \Comment{initial visual features}
\State $\mathbf{A} \leftarrow []$ \Comment{gathered cross-attention maps}
\For{$t \leftarrow 0$ \textbf{to} $T-1$}
    \State $\epsilon_\emptyset, A_{\emptyset,t} \leftarrow \boldsymbol{\epsilon_\theta}(z_t,t,\emptyset)$ \Comment{unconditional path}
    \State $\epsilon_p, A_{p,t} \leftarrow \boldsymbol{\epsilon_\theta}(z_t,t,p)$ \Comment{conditional path}
    \State $\mathbf{A} \leftarrow [\mathbf{A}, A_{p,t}]$ \Comment{add current map to the stack}
    \State $z_{t+1}\leftarrow \sqrt{\alpha_{t+1}} \left( \frac{z_t - \sqrt{1 - \alpha_t} \epsilon_\emptyset}{\sqrt{\alpha_t}} \right) + \sqrt{1 - \alpha_{t+1}} \epsilon_\emptyset$ 
    
    \State $z_t \leftarrow z_{t+1}$ \Comment{update latents}
\EndFor
\State \Return $\mathbf{A}$
\end{algorithmic}
\end{algorithm}

Using Alg. \ref{alg:ddim_inversion}, we gather a total of $N \times T$ cross-attention maps $\mathbf{A}$. However, inspired by a prior work on semantic segmentation with diffusion models \cite{baranchuk2021label}, we aim to optimise phrase grounding performance by selecting attention maps only from \textit{middle timesteps} and \textit{middle cross-attention layers}. This choice is motivated by the following two observations. First, for $t \rightarrow T$, latents $z_t$ are highly noisy and lose information about the image's global structure, thus attention maps do not convey any semantic meaning. On the contrary, for $t \rightarrow 0$, $z_t$ contains fine-grained details about the input which, according to \cite{baranchuk2021label}, hurt segmentation performance. Therefore, focusing on middle timesteps strikes a balance. Correspondingly, earlier (resp.\ final) layers are more informative for small (resp.\ large) objects in the image, meaning that those features would either activate over the entire image or only on large anatomical structures. Therefore, selecting the middle cross-attention layers balances between these two extreme cases.
% layers focus on small objects in the image, meaning that the output heatmap would highly activate over the entire image area. Conversely, final layers are more informative for larger objects such as entire anatomical structures. Therefore, focusing on middle cross-attention layers balances between these two extreme cases.

The selected cross-attention maps are intensity-normalised to span the range [0,1] and resized to match the image's spatial dimensions with bilinear interpolation. The output heatmap $\mathbf{h}$ is formed by averaging maps across layers, timesteps and tokens. Denoting the selected cross-attention layers, timesteps, and all tokens excluding padding as $L', T', S'$, respectively, Eq.\ \eqref{eq:agg} outlines this operation.

\begin{equation}
\begin{aligned}
    \mathbf{h} = \mathop{\mathbb{E}}_{l \in L'} \mathop{\mathbb{E}}_{t \in T'} \mathop{\mathbb{E}}_{s \in S'} \text{Resize}\left( \text{Norm}\left( \mathbf{A}_{s, \cdot, \cdot, l, t} \right) \right)
    %\frac{1}{|L'| |T'| |S'|} \sum^{L'} \sum^{T'} \sum^{S'} \text{Resize}\left( \text{Norm}\left( \mathbf{A}_{s, \cdot, \cdot, l, t} \right) \right)
\end{aligned}
\label{eq:agg}
\end{equation}

As a final step, results are refined by applying additional post-processing techniques on the generated heatmap $\mathbf{h}$ that do not involve any learnable parameters or any form of supervision, thus not violating the zero-shot scenario. Here, we use binary-Otsu thresholding \cite{liao2001fast} to separate the more strongly activating foreground from the more weakly activating background. The output binary mask is then applied to the heatmap $\mathbf{h}$ to suppress weak signals while leaving the foreground activations unaffected.

\section{Experimental Results}
\label{sec:results}
\subsection{LDM Pre-training Setup}
First, we briefly discuss the LDM pre-training stage. Here, we use an open-source implementation of the LDM \cite{pinaya2022brain} along with a provided checkpoint. The model is pre-trained on MIMIC-CXR \cite{johnson2019mimic}, a large-scale dataset of chest radiographs accompanied by free-text reports. The training set consists of 368,960 chest X-ray images, whereas the text prompt for each image is a randomly sampled sentence from the corresponding radiology report, from either the \textit{Impressions} or \textit{Findings} section. Note that the pre-trained weights, as well as the configuration files with all training-related details, are publicly available in the MONAI Generative framework \cite{pinaya2023generative}.

\subsection{LDM Phrase Grounding Setup}
We now instantiate our best performing setup for the phrase grounding task. We transform the input image into fixed spatial dimensions of $512\times 512$. For the diffusion process, we set the total number of timesteps of the inverse DDIM scheduler to T $=300$. In terms of text processing, we use the standard CLIP text encoder with frozen weights. Prior to encoding, the input prompt is tokenised and then either padded or truncated to match the maximum sequence length ($S=77$ tokens). Since the CLIP text encoder is not trained on radiology reports, domain-specific medical terms are out-of-vocabulary (OOV); in practice this includes all of the pathology names except ``pneumonia''. The CLIP tokeniser handles any unknown OOV word by splitting it into multiple known sub-tokens. 

The initial size of the gathered attention maps is $\mathbb{R}^{S\times B\times B\times L\times T}$, where $L=11$ is the total number of cross-attention layers in the U-Net and $B \in \{16, 32\}$ denotes their spatial size (depending on the layer's level in the U-Net). After the selection stage described in Section \ref{sec:ldmgrounding}, we end up with attention maps collected from $L=4$ layers (i.e., from the $3^{\text{rd}}$, $4^{\text{th}}$, $6^{\text{th}}$ and $7^{\text{th}}$ cross-attention layers), which automatically sets their spatial size to $B=16$, and also from 60 different timesteps (i.e., from timestep 120 to 180) out of $T=300$ steps in total. These choices are justified via the ablation study in Subsection \ref{subsec:ablation}. Note, however, that we did not perform an exhaustive search to find the optimal combination of selected layers and timesteps; in fact, it is possible that different settings per pathology might lead to better performance (we leave this for future work). Last, following \cite{boecking2022making,bannur2023learning}, the resulting heatmap $\mathbf{h}$ with resolution $512\times 512$ is convolved with a Gaussian kernel ($\sigma=2.5$) prior to Otsu thresholding.

\subsection{Evaluation Dataset}
We evaluate our proposed system on the MS-CXR benchmark \cite{boecking2022making} which consists of 1,158 image-sentence pairs with ground truth bounding boxes indicating the pathology. Note that this dataset is extracted from the official MIMIC-CXR test set. We further pre-process the dataset by merging entries corresponding to the same patient and the same sentence, i.e., an image might have more than one ground truth bounding box as reference for a given text prompt, where the pathology appears in multiple locations. We made the choice to compute metrics on a per-image basis (and not per bounding box) as this matched the performance reported by \cite{boecking2022making,bannur2023learning}.

\subsection{Baselines}
We compare our proposed method to 4 state-of-the-art baselines trained in a discriminative manner, using either fully supervised or self-supervised learning. Fully supervised networks (\textit{MedRPG} \cite{chen2023medical}, \textit{OmniFM-DR} \cite{xu2023learning}) are trained to correctly predict the ground truth bounding box, whereas the self-supervised networks (\textit{BioViL} \cite{boecking2022making}, \textit{BioViL-T} \cite{bannur2023learning}) are trained to maximise the cosine similarity between visual and textual features derived from the corresponding image and report. We evaluate our system against these 4 methods only since they were shown to outperform previous approaches, with \textit{BioViL-T} \cite{bannur2023learning} setting the current state-of-the-art on phrase grounding. For all methods, we use the publicly available pre-trained model checkpoints provided by the respective authors.

\subsection{Metrics}
Given a predicted heatmap $H$ (with values in $[0,1]$ range) and ground truth binary segmentation mask $M_{GT}$ which has ones within each bounding box area and zero otherwise, phrase grounding performance is measured via the following evaluation metrics:

\subsubsection{Mean Intersection over Union (mIoU)}
mIoU is a standard metric to evaluate segmentation performance. The predicted binary mask at threshold $thr$ is defined as $M_H = \{h\in H: h > thr\}$. Then, IoU at given threshold $thr$ is calculated as 
\begin{equation}
    IoU_{@thr} = \frac{|M_H \cap M_{GT}|}{|M_H \cup M_{GT}|}.
    \label{eq:miou}
\end{equation}
Here, following \cite{boecking2022making}, we calculate mIoU as the average over 5 different thresholds $thr\in [0.1, 0.2, 0.3, 0.4, 0.5]$.

\subsubsection{Area Under ROC Curve (AUC-ROC)}
The area under the receiver operating characteristic curve is another method to estimate segmentation performance on a per-pixel basis given $H$ and $M_{GT}$.

\subsubsection{Contrast-to-Noise Ratio (CNR)}
CNR as used in \cite{boecking2022making} is a threshold-agnostic measure that reflects the distribution of raw heatmap activations over the entire input. Let $A$ denote the area within each bounding box, whereas $\bar{A}$ is the rest of the heatmap, i.e., $H=A\cup \bar{A}$. Then, after calculating the mean $\mu$ and variance $\sigma^2$ of the raw heatmap scores for each area $A$ and $\bar{A}$, respectively, CNR is defined as
\begin{equation}
    CNR = \frac{\mu_A - \mu_{\bar{A}}}{\sqrt{\sigma_A^2 + \sigma_{\bar{A}}^2}}.
    \label{eq:cnr}
\end{equation}
Note that this definition will penalise the case where $\mu_A < \mu_{\bar{A}}$. However, following \cite{boecking2022making}, we also provide results for the absolute CNR evaluated as 
\begin{equation}
    |CNR| = \frac{|\mu_A - \mu_{\bar{A}}|}{\sqrt{\sigma_A^2 + \sigma_{\bar{A}}^2}}.
    \label{eq:cnr_abs}
\end{equation}

\subsection{Evaluation Protocol}
To ensure a fair comparison across all approaches, we adopt the following protocol:
\begin{itemize}
    \item All metrics are computed on the original image dimensions. Therefore, for heatmap-based methods such as \cite{boecking2022making,bannur2023learning}, and our own method, we first perform nearest neighbor interpolation with appropriate zero padding to match the original image resolution. Note that each of those methods initially generates a fixed size heatmap which matches the respective input size: for BioViL \cite{boecking2022making} the resolution is $480\times 480$, for BioViL-T \cite{bannur2023learning} it is equal to $448\times 448$, while our LDM-based approach outputs a $512\times 512$ heatmap per image. For methods \cite{chen2023medical} and \cite{xu2023learning} that predict bounding box coordinates, the predicted bounding box regions are resized to the width and height of the original image.
    \item Since heatmaps generated with \textit{BioViL} \cite{boecking2022making} and \textit{BioViL-T} \cite{bannur2023learning} methods are in the $[-1,1]$ range (due to cosine similarity), whereas our method yields heatmaps in the $[0,1]$ range, we set all negative values for the aforementioned baselines \cite{boecking2022making,bannur2023learning} to 0.
    \item Both \textit{MedRPG} \cite{chen2023medical} and \textit{OmniFM-DR} \cite{xu2023learning} are Transformer models that predict bounding box coordinates. Thus, we evaluate \cite{chen2023medical,xu2023learning} only in terms of mIoU.
    \item \textit{MedRPG} \cite{chen2023medical} does not support the case where more than one bounding box exists per image. Therefore, for those images, we evaluate it on each bounding box separately and only report the maximum value of mIoU per image.
\end{itemize}

\subsection{Results and Discussion}
%What is the big picture. What are the major questions to answer. How ablation identify strenght in the choosen approach.
In this section we report results using the 4 aforementioned metrics (mIoU, AUC-ROC, $|$CNR$|$ and CNR), and also discuss implications. Overall phrase grounding results on the MS-CXR database are reported in Table \ref{tab1}. In Subsection \ref{subsec:ablation} we show the results of a separate ablation study.

\begin{table*}[ht]
\centering
\caption{Phrase grounding results on MS-CXR dataset. Mean results, and their respective standard deviation, are reported across the 8 pathologies of interest and also averaged (Avg). N denotes the sample size per pathology. Both $|$CNR$|$ and CNR metrics are unbounded, i.e., in $(-\infty , \infty)$ range. mIoU and AUC-ROC are reported in (\%). Higher is better for all metrics ($\uparrow$). Note that, due to implementation details, only mIoU can be computed for methods \cite{chen2023medical} and \cite{xu2023learning}. Best metrics are highlighted with \textbf{bold}. Second best metrics are \underline{underlined}.}
\label{table}
\setlength{\tabcolsep}{3pt}
\renewcommand{\arraystretch}{1.2}
%\begin{tabular}{|p{25pt}|p{75pt}|p{115pt}|}
\begin{tabular}{|c|c|c|c|c|c|c|c|c|c|c|}
\hline
\multirow{2}{*}{\textbf{Method}} &
\multirow{2}{*}{\textbf{Metric}} &
\textbf{Pneumonia} &
\textbf{Pneumothorax} &
\textbf{Consolidation} &
\textbf{Atelectasis} &
\textbf{Edema} &
\textbf{Cardiomeg.} &
\textbf{Lung Opac.} &
\textbf{Pleural Eff.} &
\multirow{2}{*}{\textbf{Avg}} \\
& & (N=182) & (N=243) & (N=117) & (N=61) & (N=44) & (N=333) & (N=82) & (N=96) & \\
\hline 
MedRPG \cite{chen2023medical}& mIoU& 11.3& 5.8& 6.1& 12.2& 5.1& 16.3& 6.5& 11.7& 9.4 \\  %[0.1ex]
\hline
OmniFM-DR \cite{xu2023learning}& mIoU& 12.6& 7.0& 24.7& 5.4& 17.0& \underline{30.9}& 15.4& 5.1& 14.8 \\  %[0.1ex]
\hline
\multirow{4}{*}{BioViL \cite{boecking2022making}}& $|$CNR$|$& \underline{1.56}$_{\, 0.80}$& \underline{0.78}$_{\, 0.56}$& \textbf{1.79}$_{\, 0.77}$& \underline{1.37}$_{\, 0.65}$& \textbf{0.85}$_{\, 0.57}$& 0.81$_{\, 0.54}$& \underline{1.24}$_{\, 0.81}$& \underline{1.38}$_{\, 0.71}$& \underline{1.22}$_{\, 0.35}$ \\
& CNR& \underline{1.49}$\, _{0.89}$ & \underline{0.63}$\, _{0.71}$ & \underline{1.73}$\, _{0.83}$ & \underline{1.28}$\, _{0.77}$ & \textbf{0.77}$\, _{0.60}$ & 0.73$\, _{0.64}$ & \underline{1.18}$\, _{0.88}$ & \underline{1.33}$\, _{0.78}$ & \underline{1.14}$\, _{0.37}$ \\
& mIoU& \underline{27.3}$_{\, 16.0}$ & \underline{10.2}$_{\, 10.2}$ & \textbf{31.8}$_{\, 14.3}$ & 24.1$_{\, 14.2}$ & \underline{21.3}$_{\, 17.2}$ & 22.0$_{\, 18.6}$ & 15.0$_{\, 14.1}$ & \textbf{20.4}$_{\, 11.6}$ & 21.5$_{\, 6.31}$ \\
& AUC-ROC& 76.5$_{\, 16.4}$ & \underline{66.3}$_{\, 17.5}$ & \underline{83.5}$_{\, 12.1}$ & \underline{76.4}$_{\, 15.1}$ & \underline{65.1}$_{\, 13.1}$ & 64.5$_{\, 14.3}$ & 68.9$_{\, 17.2}$ & \textbf{76.6}$_{\, 15.8}$ & 72.2$_{\, 6.50}$ \\
\hline
\multirow{4}{*}{BioViL-T \cite{bannur2023learning}}& $|$CNR$|$& \textbf{1.70}$_{\, 0.70}$& \textbf{1.01}$_{\, 0.63}$& \textbf{1.79}$_{\, 0.76}$& \textbf{1.47}$_{\, 0.66}$& \underline{0.84}$_{\, 0.46}$& \textbf{1.06}$_{\, 0.50}$& \textbf{1.59}$_{\, 0.87}$& \textbf{1.55}$_{\, 0.67}$& \textbf{1.38}$_{\, 0.33}$\\
& CNR& \textbf{1.66}$_{\, 0.77}$& \textbf{0.91}$_{\, 0.75}$& \textbf{1.74}$_{\, 0.82}$& \textbf{1.45}$_{\, 0.69}$& \textbf{0.77}$_{\, 0.50}$& \textbf{1.05}$_{\, 0.52}$& \textbf{1.54}$_{\, 0.94}$& \textbf{1.53}$_{\, 0.70}$& \textbf{1.33}$_{\, 0.34}$\\
& mIoU& \textbf{29.0}$_{\, 13.9}$ & \textbf{12.6}$_{\, 12.6}$ & \underline{30.2}$_{\, 12.9}$ & \underline{24.7}$_{\, 12.8}$ & 19.0$_{\, 15.1}$ & 23.5$_{\, 15.3}$ & \textbf{17.5}$_{\, 12.6}$ & 18.6$_{\, 10.2}$ & \underline{21.9}$_{\, 5.65}$ \\
& AUC-ROC& \textbf{80.1}$_{\, 15.0}$ & \textbf{70.1}$_{\, 18.0}$ & \textbf{83.9}$_{\, 11.7}$ & \underline{76.4}$_{\, 14.4}$ & 63.0$_{\, 11.9}$ & \underline{66.1}$_{\, 12.6}$ & \underline{76.3}$_{\, 16.1}$ & 73.8$_{\, 15.0}$ & \underline{73.7}$_{\, 6.58}$ \\
\hline
\multirow{4}{*}{\textbf{Ours}}& $|$CNR$|$& 1.02$_{\, 0.46}$ & 0.46$_{\, 0.31}$ & 1.18$_{\, 0.55}$ & 1.06$_{\, 0.50}$ & 0.76$_{\, 0.36}$ & \underline{0.91}$_{\, 0.43}$ & 1.12$_{\, 0.57}$ & 0.88$_{\, 0.43}$ & 0.92$_{\, 0.22}$ \\
& CNR& 1.02$_{\, 0.47}$ & -0.08$_{\, 0.53}$ & 1.16$_{\, 0.57}$ & 1.06$_{\, 0.51}$ & 0.71$_{\, 0.45}$ & \underline{0.90}$_{\, 0.44}$ & 1.08$_{\, 0.63}$ & 0.86$_{\, 0.46}$ & 0.84$_{\, 0.37}$ \\
& mIoU& 23.8$_{\, 11.9}$ & 5.30$_{\, 5.30}$ & 24.7$_{\, 14.4}$ & \textbf{25.0}$_{\, 11.1}$ & \textbf{30.5}$_{\, 19.6}$ & \textbf{37.4}$_{\, 11.6}$ & \underline{16.7}$_{\, 13.0}$ & \underline{19.3}$_{\, 9.93}$ & \textbf{22.8}$_{\, 8.94}$ \\
& AUC-ROC& \underline{78.9}$_{\, 9.78}$ & 49.5$_{\, 14.8}$ & 81.1$_{\, 9.98}$ & \textbf{79.1}$_{\, 10.1}$ & \textbf{72.4}$_{\, 11.4}$ & \textbf{75.3}$_{\, 10.0}$ & \textbf{79.6}$_{\, 13.0}$ & \underline{75.2}$_{\, 10.7}$ & \textbf{73.9}$_{\, 9.62}$ \\
\hline
\end{tabular}
\label{tab1}
\end{table*}

%Overall phrase grounding results on the MS-CXR database are reported in Table \ref{tab1}.  \todo{{\footnotesize R3.5}}\textcolor{red}{Note that error bars are not included in our analysis since all models were run on inference mode, thus yielding deterministic outputs.}

We now provide an interpretation of the results shown in Table \ref{tab1}. More specifically, we draw the following conclusions: First, our proposed method outperforms both supervised baselines MedRPG \cite{chen2023medical} and OmniFM-DR \cite{xu2023learning} by a large margin. This also holds for the baselines \cite{boecking2022making} and \cite{bannur2023learning} trained on image-text pairs via self-supervision. This suggests that phrase grounding performance is largely affected by the size of the pre-training dataset, while bounding box annotations are typically scarce. 
% free-text reports provide a rich source of information that can be used as an alternative to ground truth bounding boxes\textcolor{red}{, especially when those annotations are scarce}

Second, our method is competitive with both BioViL variants \cite{boecking2022making,bannur2023learning} for most pathologies on the MS-CXR dataset. In fact, our pipeline based on the LDM sets a new state-of-the-art in terms of both mIoU (0.9 $\%$ relative improvement to BioViL-T \cite{bannur2023learning}) and AUC-ROC (0.2 $\%$ relative increase to BioViL-T \cite{bannur2023learning}) metrics averaged across all classes. Note also that both BioViL models \cite{boecking2022making,bannur2023learning} use radiology-specific text encoders which are expected to further improve performance, whereas our method relies on a frozen CLIP text encoder pre-trained on data collected from the Internet. In addition, unlike all other approaches that use discriminative models such as ResNets \cite{boecking2022making,bannur2023learning} (or Transformers \cite{xu2023learning,chen2023medical}), the LDM is based on a U-Net for feature extraction, thus its representations are readily applicable to localisation tasks such as phrase grounding. We also observe that the results exhibit high variance across all methods. This is likely a data related issue and needs to be further investigated in the future.

Furthermore, given both definitions of the CNR metric presented in Eqs.\ \eqref{eq:cnr} and \eqref{eq:cnr_abs}, respectively, our method remains fairly robust between the two. Specifically, our approach yields the lowest difference between $|$CNR$|$ and CNR for 7 out of 8 pathologies. This highlights that $|$CNR$|$, which is defined in \cite{boecking2022making} and \cite{bannur2023learning}, overestimates performance, thus it is less reliable. 

We also observe that every method performs poorly on the \textit{Pneumothorax} class (our approach leads to a negative CNR value). We
note that pneumothorax causes a dark air space (i.e., a region with low intensity pixels) in the position of the collapsed lung, as opposed to the other pathologies which manifest as ``bright'' regions. These dark regions may be more difficult to differentiate from normal lung.
%, and since dark areas tend to be generated by an \emph{absence} of activations in diffusion models, our proposed localisation approach may fail.
%that it is more difficult to localise accurately, especially with our proposed approach.

\subsection{Ablation Study}
\label{subsec:ablation}
% Note that, although we got the exact same results for T=40, we chose T=20 since it speeds up inference. 
In Table \ref{tab2} we provide an ablation study showing how different hyperparameters affect phrase grounding performance. To speed up experiments, we set the total number of timesteps to T=100, yet the empirical observations are expected to also hold for larger T. Starting from the initial setup of collecting all attention maps (L=11, T=100), we focus either on middle cross-attention layers (L=6) or middle timesteps (T=20). We see that both of the aforementioned choices have a positive impact on both metrics. Furthermore, we observe that the combination thereof (L=6, T=20) yields a substantial boost in performance compared to the initial setup (L=11, T=100). We also show the effect of using binary Otsu thresholding. In fact, although we notice a slight decrease in CNR, mIoU is increased by 3.5$\%$. Note also how different pathologies might benefit from different setups.

Moreover, the ablation study presented in Table \ref{tab3} shows the impact of prompt tokens on phrase grounding performance. To this end, while fixing all other components of our system (L=6, T=20 out of 100 timesteps in total), we discard cross-attention maps that are not related to pathology tokens i.e., the (sub-)tokens corresponding to the pathology label names. Also, for this experiment, we filtered out inputs that do not contain the pathology name in the prompt, thus reducing the dataset to 694 image-text pairs. The results in Table \ref{tab3} indicate that cross-attention maps corresponding to pathology tokens are not sufficient to perform phrase grounding (interestingly, when we only use pathology tokens, performance on \textit{Pneumothorax} class slightly improves; however, this approach clearly underperforms on all other classes). This can be attributed to the fact that text prompts usually contain additional important information such as location (e.g., \textit{right}, \textit{left}, \textit{bibasilar}) and severity (e.g., \textit{mild}, \textit{moderate}, \textit{severe}) modifiers that can be used to localise the underlying pathology.

\begin{table*}[ht]
\centering
\caption{Ablation study on the choice of layers L and timesteps T, as well as the effect of applying Otsu thresholding. Note that T=20 refers to the $[40,60]$ range out of 100 steps in total, while L=6 includes layers with index in $\{3, ..., 8\}$ range out of 11 in total. CNR metric is unbounded, mIoU is in ($\%$). Best metrics are highlighted with \textbf{bold}.}
\label{table2}
\setlength{\tabcolsep}{3pt}
\renewcommand{\arraystretch}{1.2}
%\begin{tabular}{|p{25pt}|p{75pt}|p{115pt}|}
\begin{tabular}{|c|c|c|c|c|c|c|c|c|c|c|}
\hline
\textbf{Setup}&
\textbf{Metric}&
\textbf{Pneumonia}&
\textbf{Pneumothorax}&
\textbf{Consolidation}&
\textbf{Atelectasis}&
\textbf{Edema}&
\textbf{Cardiomegaly}&
\textbf{Lung Opacity}&
\textbf{Pleural Eff.}&
\textbf{Avg} \\ 
\hline
\multirow{2}{*}{L=11, T=100}& CNR& 0.87 & -0.15 & 0.90 & 0.98 & 0.56 & 0.52 & 0.90 & \textbf{0.89} & 0.68 \\
& mIoU& 15.0 & 5.4 & 16.3 & 16.5 & 25.2 & 22.9 & 10.8 & 13.9 & 15.7 \\
\hline
\multirow{2}{*}{L=11, T=20}& CNR& 0.89 & -0.15 & 0.94 & 0.97 & 0.66 & 0.60 & 0.91 & 0.88 & 0.71 \\
& mIoU& 16.1 & 5.3 & 17.3 & 17.2 & 25.7 & 24.3 & 11.8 & 14.0 & 16.5 \\
\hline 
\multirow{2}{*}{L=6, T=100}& CNR& 0.89 & -0.08 & 0.98 & \textbf{1.03} & 0.58 & 0.72 & 0.98 & \textbf{0.89} & 0.75 \\
& mIoU& 15.7 & 5.5 & 17.0 & 17.2 & 25.7 & 25.7 & 11.1 & 14.4 & 16.6 \\
\hline 
\multirow{2}{*}{L=6, T=20}& CNR& \textbf{0.92} & \textbf{-0.05} & \textbf{1.02} & \textbf{1.03} & \textbf{0.67} & 0.79 & \textbf{0.99} & \textbf{0.89} & \textbf{0.78} \\
& mIoU& 15.8 & \textbf{5.7} & 17.4 & 17.0 & 26.8 & 27.5 & 11.6 & 13.9 & 17.0 \\
\hline
% \multirow{2}{*}{L=6, T=40, Otsu}& CNR& \textbf{0.87} & -0.16 & \textbf{0.98} & \textbf{0.99} & 0.56 & 0.78 & \textbf{0.94} & 0.87 & 0.73 \\
% & mIoU& 20.2 & 5.0 & \textbf{21.3} & \textbf{22.1} & 27.6 & 32.4 & 13.9 & \textbf{18.1} & 20.1 \\
% \hline 
\multirow{2}{*}{L=6, T=20, Otsu}& CNR& 0.89 & -0.14 & 1.00 & 1.00 & 0.62 & \textbf{0.81} & 0.96 & 0.85 & 0.75 \\
& mIoU& \textbf{20.5} & 5.1 & \textbf{21.4} & \textbf{22.2} & \textbf{28.5} & \textbf{33.3} & \textbf{14.4} & \textbf{17.8} & \textbf{20.4} \\
\hline 
\end{tabular}
\label{tab2}
\end{table*}

\begin{table*}[ht]
\centering
\caption{Ablation study on the effect of the selected prompt tokens. In the first setup (\textit{Pathology tokens}), cross-attention maps are extracted only from tokens related to the pathology, whereas the second setup (\textit{All tokens}) considers the entire prompt. CNR metric is unbounded, mIoU is in ($\%$). Best metrics are highlighted with \textbf{bold}.}
\label{table3}
\setlength{\tabcolsep}{3pt}
\renewcommand{\arraystretch}{1.2}
%\begin{tabular}{|p{25pt}|p{75pt}|p{115pt}|}
\begin{tabular}{|c|c|c|c|c|c|c|c|c|c|c|}
\hline
\textbf{Setup}&
\textbf{Metric}&
\textbf{Pneumonia}&
\textbf{Pneumothorax}&
\textbf{Consolidation}&
\textbf{Atelectasis}&
\textbf{Edema}&
\textbf{Cardiomegaly}&
\textbf{Lung Opacity}&
\textbf{Pleural Eff.}&
\textbf{Avg} \\ 
\hline
\multirow{2}{*}{Pathology tokens}& CNR& 0.05 & \textbf{0.25} & 0.40 & 0.39 & 0.45 & 0.41 & 0.00 & 0.50 & 0.31 \\
& mIoU& 9.50 & \textbf{6.76} & 14.7 & 13.3 & 22.4 & 25.9 & 2.59 & 13.4 & 13.6 \\
\hline
\multirow{2}{*}{All tokens}& CNR& \textbf{0.93} & -0.11 & \textbf{1.12} & \textbf{1.04} & \textbf{0.64} & \textbf{0.96} & \textbf{0.59} & \textbf{0.89} & \textbf{0.76} \\
& mIoU& \textbf{21.8} & 5.07 & \textbf{21.9} & \textbf{22.9} & \textbf{29.5} & \textbf{41.4} & \textbf{6.55} & \textbf{17.8} & \textbf{20.9} \\
\hline 
\end{tabular}
\label{tab3}
\end{table*}

\subsection{Qualitative Analysis}
\label{subsec:qualitative}
\begin{figure*}[!t]
\centering
\includegraphics[width=\textwidth,height=12.2cm,keepaspectratio]{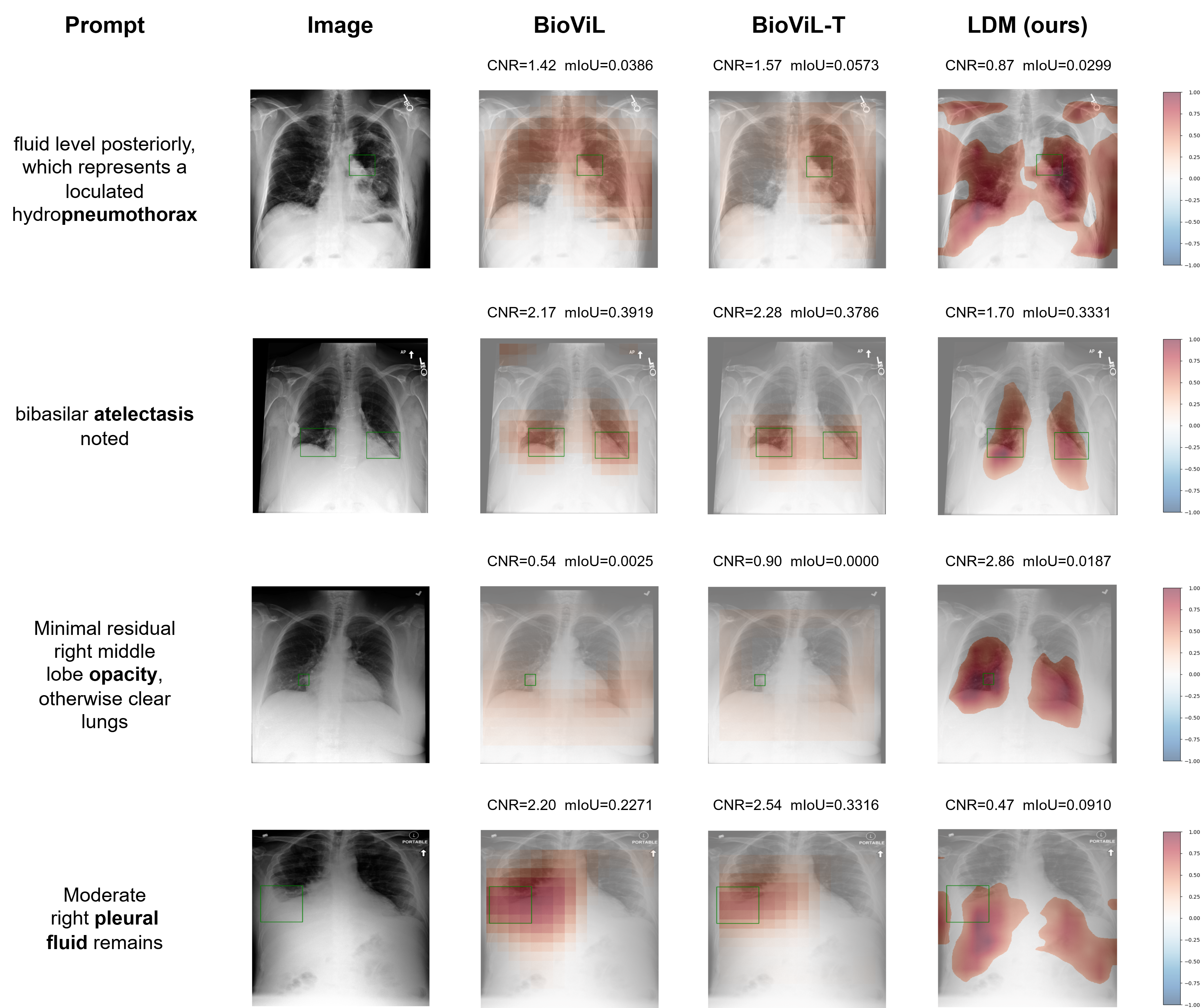}  %height=12.5cm
\caption{Randomly selected results for the phrase grounding task. For each input image-prompt pair, we show the heatmaps generated from \textit{BioViL} \cite{boecking2022making}, \textit{BioViL-T} \cite{bannur2023learning} and our own method, respectively, overlaid on the original images. Ground truth classes are highlighted in \textbf{bold} within each prompt. Ground truth bounding boxes are depicted in green. For each method, we also provide the reported $|$CNR$|$ and mIoU metrics (shown on top of each figure). Best viewed in colour.}
\label{fig3}
\end{figure*}

Fig.\ \ref{fig3} depicts non cherry-picked examples from the MS-CXR dataset and the generated heatmaps for BioViL \cite{boecking2022making}, BioViL-T \cite{bannur2023learning}, and our method. We observe that both BioViL models provide more densely localised results compared to our system, which activates on larger input areas. However, unlike BioViL(-T), our method does not miss an area of interest (cf.\ third row of Fig.\ \ref{fig3} for pathology \textit{Lung Opacity}). We note that our LDM-based method could also be focusing on less relevant anatomical regions in some cases (cf.\ first row of Fig.\ \ref{fig3} for \textit{Pneumothorax}) that can be recognised as easy false positives.

\section{Conclusion}
\label{sec:conclusion}
In this work we have presented a novel approach for performing phrase grounding with a pre-trained Latent Diffusion Model. In fact, we draw on the parts of the model that integrate visual and textual features, namely the cross-attention layers. These layers, as evidenced by our results, provide a rich source of information that can be readily used to solve the task at hand. Our proposed method does not alter the backbone generative model in any way, thus operating in zero-shot. 

Our proposed system is limited by the computational cost of the LDM sampling process (one inference per timestep), which leads to slower inference speed compared to other baselines. We also identified certain pathologies (e.g., \textit{Pneumothorax}) where all models, including ours, underperform; this requires further analysis as it might indicate hidden data biases. We also expect that an LDM trained on chest X-rays will perform poorly in significantly different medical contexts (e.g., brain MRIs) without fine-tuning.

Regarding future work, it is worth experimenting with few-shot fine-tuning methods (e.g., low rank adaptation \cite{hu2021lora}) that would allow us to incorporate new knowledge to the pre-trained LDM, or even adapt it to unknown data distributions, with a small target dataset. In this direction, according to prior works in zero-shot domain adaptation, we would likely require access to either task-irrelevant target data \cite{peng2018zero} or target domain-specific prompts \cite{fahes2023poda,yang2024unified} to make our zero-shot method robust to domain shifts. Moreover, devising faster sampling methods tailored for the phrase grounding task would render the LDM more efficient for real-world scenarios. We also believe that further improvements on the generative aspect of the LDM (e.g., mitigating various data biases \cite{perez2023radedit}) will bring a positive effect on the model's downstream performance.

In terms of broader impact, our proposed framework might be used to automatically link reports to the relevant image locations, allowing fast inclusion of key images and easy navigation when reviewing a previous scan. We might also extend to the task of diagnosis by creating text prompts such as \texttt{"Where is \{pathology\_label\}?"}, to achieve an off-the-shelf detector (beyond the scope of this paper).
%  that provides automated diagnosis
%our proposed framework can be easily extended to enable real-world applicability. Specifically, it is clear that our method cannot be directly used for real-world diagnosis, since text information is only available after a doctor provides an interpretation of the scan. However, we can alleviate this issue by generating synthetic text prompts such as \texttt{"Where is \{pathology\_label\}?"}, thus ending up with an off-the-shelf detector that provides an automated diagnosis.

% \appendices

% Appendixes, if needed, appear before the acknowledgment.

%\bibliographystyle{IEEEtran}
%\bibliography{bibliography.bib} 
\printbibliography

@inproceedings{boecking2022making,
  title={Making the most of text semantics to improve biomedical vision--language processing},
  author={Boecking, Benedikt and Usuyama, Naoto and Bannur, Shruthi and Castro, Daniel C and Schwaighofer, Anton and Hyland, Stephanie and Wetscherek, Maria and Naumann, Tristan and Nori, Aditya and Alvarez-Valle, Javier and others},
  booktitle={European conference on computer vision},
  pages={1--21},
  year={2022},
  organization={Springer}
}

@article{dhariwal2021diffusion,
  title={Diffusion models beat gans on image synthesis},
  author={Dhariwal, Prafulla and Nichol, Alexander},
  journal={Advances in neural information processing systems},
  volume={34},
  pages={8780--8794},
  year={2021}
}

@inproceedings{bannur2023learning,
  title={Learning to exploit temporal structure for biomedical vision-language processing},
  author={Bannur, Shruthi and Hyland, Stephanie and Liu, Qianchu and Perez-Garcia, Fernando and Ilse, Maximilian and Castro, Daniel C and Boecking, Benedikt and Sharma, Harshita and Bouzid, Kenza and Thieme, Anja and others},
  booktitle={Proceedings of the IEEE/CVF Conference on Computer Vision and Pattern Recognition},
  pages={15016--15027},
  year={2023}
}

@misc{dombrowski2023tradeoffs,
      title={Trade-offs in Fine-tuned Diffusion Models Between Accuracy and Interpretability}, 
      author={Mischa Dombrowski and Hadrien Reynaud and Johanna P. Müller and Matthew Baugh and Bernhard Kainz},
      year={2023},
      eprint={2303.17908},
      archivePrefix={arXiv},
      primaryClass={cs.CV}
}

@inproceedings{chen2023medical,
  title={Medical phrase grounding with region-phrase context contrastive alignment},
  author={Chen, Zhihao and Zhou, Yang and Tran, Anh and Zhao, Junting and Wan, Liang and Ooi, Gideon Su Kai and Cheng, Lionel Tim-Ee and Thng, Choon Hua and Xu, Xinxing and Liu, Yong and others},
  booktitle={International Conference on Medical Image Computing and Computer-Assisted Intervention},
  pages={371--381},
  year={2023},
  organization={Springer}
}

@article{xu2023learning,
  title={Learning A Multi-Task Transformer Via Unified And Customized Instruction Tuning For Chest Radiograph Interpretation},
  author={Xu, Lijian and Ni, Ziyu and Liu, Xinglong and Wang, Xiaosong and Li, Hongsheng and Zhang, Shaoting},
  journal={arXiv preprint arXiv:2311.01092},
  year={2023}
}

@article{dawidowicz2023limitr,
  title={LIMITR: Leveraging Local Information for Medical Image-Text Representation},
  author={Dawidowicz, Gefen and Hirsch, Elad and Tal, Ayellet},
  journal={arXiv preprint arXiv:2303.11755},
  year={2023}
}

@article{pinaya2023generative,
  title={Generative ai for medical imaging: extending the monai framework},
  author={Pinaya, Walter HL and Graham, Mark S and Kerfoot, Eric and Tudosiu, Petru-Daniel and Dafflon, Jessica and Fernandez, Virginia and Sanchez, Pedro and Wolleb, Julia and da Costa, Pedro F and Patel, Ashay and others},
  journal={arXiv preprint arXiv:2307.15208},
  year={2023}
}

@inproceedings{pinaya2022brain,
  title={Brain imaging generation with latent diffusion models},
  author={Pinaya, Walter HL and Tudosiu, Petru-Daniel and Dafflon, Jessica and Da Costa, Pedro F and Fernandez, Virginia and Nachev, Parashkev and Ourselin, Sebastien and Cardoso, M Jorge},
  booktitle={MICCAI Workshop on Deep Generative Models},
  pages={117--126},
  year={2022},
  organization={Springer}
}

@inproceedings{rombach2022high,
  title={High-resolution image synthesis with latent diffusion models},
  author={Rombach, Robin and Blattmann, Andreas and Lorenz, Dominik and Esser, Patrick and Ommer, Bj{\"o}rn},
  booktitle={Proceedings of the IEEE/CVF conference on computer vision and pattern recognition},
  pages={10684--10695},
  year={2022}
}

@inproceedings{mokady2023null,
  title={Null-text inversion for editing real images using guided diffusion models},
  author={Mokady, Ron and Hertz, Amir and Aberman, Kfir and Pritch, Yael and Cohen-Or, Daniel},
  booktitle={Proceedings of the IEEE/CVF Conference on Computer Vision and Pattern Recognition},
  pages={6038--6047},
  year={2023}
}

@article{johnson2019mimic,
  title={MIMIC-CXR-JPG, a large publicly available database of labeled chest radiographs},
  author={Johnson, Alistair EW and Pollard, Tom J and Greenbaum, Nathaniel R and Lungren, Matthew P and Deng, Chih-ying and Peng, Yifan and Lu, Zhiyong and Mark, Roger G and Berkowitz, Seth J and Horng, Steven},
  journal={arXiv preprint arXiv:1901.07042},
  year={2019}
}

@article{baranchuk2021label,
  title={Label-efficient semantic segmentation with diffusion models},
  author={Baranchuk, Dmitry and Rubachev, Ivan and Voynov, Andrey and Khrulkov, Valentin and Babenko, Artem},
  journal={arXiv preprint arXiv:2112.03126},
  year={2021}
}

@article{liao2001fast,
  title={A fast algorithm for multilevel thresholding},
  author={Liao, Ping-Sung and Chen, Tse-Sheng and Chung, Pau-Choo and others},
  journal={J. Inf. Sci. Eng.},
  volume={17},
  number={5},
  pages={713--727},
  year={2001}
}

@article{lu2019vilbert,
  title={Vilbert: Pretraining task-agnostic visiolinguistic representations for vision-and-language tasks},
  author={Lu, Jiasen and Batra, Dhruv and Parikh, Devi and Lee, Stefan},
  journal={Advances in neural information processing systems},
  volume={32},
  year={2019}
}

@inproceedings{li2022grounded,
  title={Grounded language-image pre-training},
  author={Li, Liunian Harold and Zhang, Pengchuan and Zhang, Haotian and Yang, Jianwei and Li, Chunyuan and Zhong, Yiwu and Wang, Lijuan and Yuan, Lu and Zhang, Lei and Hwang, Jenq-Neng and others},
  booktitle={Proceedings of the IEEE/CVF Conference on Computer Vision and Pattern Recognition},
  pages={10965--10975},
  year={2022}
}

@inproceedings{kamath2021mdetr,
  title={Mdetr-modulated detection for end-to-end multi-modal understanding},
  author={Kamath, Aishwarya and Singh, Mannat and LeCun, Yann and Synnaeve, Gabriel and Misra, Ishan and Carion, Nicolas},
  booktitle={Proceedings of the IEEE/CVF International Conference on Computer Vision},
  pages={1780--1790},
  year={2021}
}

@article{chen2023language,
  title={Language-Guided Diffusion Model for Visual Grounding},
  author={Chen, Sijia and Li, Baochun},
  journal={arXiv preprint arXiv:2308.09599},
  year={2023}
}

@article{li2023your,
  title={Your diffusion model is secretly a zero-shot classifier},
  author={Li, Alexander C and Prabhudesai, Mihir and Duggal, Shivam and Brown, Ellis and Pathak, Deepak},
  journal={arXiv preprint arXiv:2303.16203},
  year={2023}
}

@article{clark2024text,
  title={Text-to-Image Diffusion Models are Zero Shot Classifiers},
  author={Clark, Kevin and Jaini, Priyank},
  journal={Advances in Neural Information Processing Systems},
  volume={36},
  year={2024}
}

@article{zhao2023unleashing,
  title={Unleashing text-to-image diffusion models for visual perception},
  author={Zhao, Wenliang and Rao, Yongming and Liu, Zuyan and Liu, Benlin and Zhou, Jie and Lu, Jiwen},
  journal={arXiv preprint arXiv:2303.02153},
  year={2023}
}

@inproceedings{bhalodia2021improving,
  title={Improving pneumonia localization via cross-attention on medical images and reports},
  author={Bhalodia, Riddhish and Hatamizadeh, Ali and Tam, Leo and Xu, Ziyue and Wang, Xiaosong and Turkbey, Evrim and Xu, Daguang},
  booktitle={Medical Image Computing and Computer Assisted Intervention--MICCAI 2021: 24th International Conference, Strasbourg, France, September 27--October 1, 2021, Proceedings, Part II 24},
  pages={571--581},
  year={2021},
  organization={Springer}
}

@article{bommasani2021opportunities,
  title={On the opportunities and risks of foundation models},
  author={Bommasani, Rishi and Hudson, Drew A and Adeli, Ehsan and Altman, Russ and Arora, Simran and von Arx, Sydney and Bernstein, Michael S and Bohg, Jeannette and Bosselut, Antoine and Brunskill, Emma and others},
  journal={arXiv preprint arXiv:2108.07258},
  year={2021}
}

@article{perez2023radedit,
  title={RadEdit: stress-testing biomedical vision models via diffusion image editing},
  author={P{\'e}rez-Garc{\'\i}a, Fernando and Bond-Taylor, Sam and Sanchez, Pedro P and van Breugel, Boris and Castro, Daniel C and Sharma, Harshita and Salvatelli, Valentina and Wetscherek, Maria TA and Richardson, Hannah and Lungren, Matthew P and others},
  journal={arXiv preprint arXiv:2312.12865},
  year={2023}
}

@inproceedings{ronneberger2015u,
  title={U-net: Convolutional networks for biomedical image segmentation},
  author={Ronneberger, Olaf and Fischer, Philipp and Brox, Thomas},
  booktitle={Medical image computing and computer-assisted intervention--MICCAI 2015: 18th international conference, Munich, Germany, October 5-9, 2015, proceedings, part III 18},
  pages={234--241},
  year={2015},
  organization={Springer}
}

@article{sanchez2022diffusion,
  title={Diffusion causal models for counterfactual estimation},
  author={Sanchez, Pedro and Tsaftaris, Sotirios A},
  journal={arXiv preprint arXiv:2202.10166},
  year={2022}
}

@article{bedel2023dreamr,
  title={Dreamr: Diffusion-driven counterfactual explanation for functional mri},
  author={Bedel, Hasan Atakan and {\c{C}}ukur, Tolga},
  journal={arXiv preprint arXiv:2307.09547},
  year={2023}
}

@inproceedings{wolleb2022diffusion,
  title={Diffusion models for implicit image segmentation ensembles},
  author={Wolleb, Julia and Sandk{\"u}hler, Robin and Bieder, Florentin and Valmaggia, Philippe and Cattin, Philippe C},
  booktitle={International Conference on Medical Imaging with Deep Learning},
  pages={1336--1348},
  year={2022},
  organization={PMLR}
}

@inproceedings{sanchez2022healthy,
  title={What is healthy? generative counterfactual diffusion for lesion localization},
  author={Sanchez, Pedro and Kascenas, Antanas and Liu, Xiao and O’Neil, Alison Q and Tsaftaris, Sotirios A},
  booktitle={MICCAI Workshop on Deep Generative Models},
  pages={34--44},
  year={2022},
  organization={Springer}
}

@inproceedings{pinaya2022fast,
  title={Fast unsupervised brain anomaly detection and segmentation with diffusion models},
  author={Pinaya, Walter HL and Graham, Mark S and Gray, Robert and Da Costa, Pedro F and Tudosiu, Petru-Daniel and Wright, Paul and Mah, Yee H and MacKinnon, Andrew D and Teo, James T and Jager, Rolf and others},
  booktitle={International Conference on Medical Image Computing and Computer-Assisted Intervention},
  pages={705--714},
  year={2022},
  organization={Springer}
}

@article{hertz2022prompt,
  title={Prompt-to-prompt image editing with cross attention control},
  author={Hertz, Amir and Mokady, Ron and Tenenbaum, Jay and Aberman, Kfir and Pritch, Yael and Cohen-Or, Daniel},
  journal={arXiv preprint arXiv:2208.01626},
  year={2022}
}

@article{tang2022daam,
  title={What the daam: Interpreting stable diffusion using cross attention},
  author={Tang, Raphael and Liu, Linqing and Pandey, Akshat and Jiang, Zhiying and Yang, Gefei and Kumar, Karun and Stenetorp, Pontus and Lin, Jimmy and Ture, Ferhan},
  journal={arXiv preprint arXiv:2210.04885},
  year={2022}
}

@inproceedings{patashnik2023localizing,
  title={Localizing object-level shape variations with text-to-image diffusion models},
  author={Patashnik, Or and Garibi, Daniel and Azuri, Idan and Averbuch-Elor, Hadar and Cohen-Or, Daniel},
  booktitle={Proceedings of the IEEE/CVF International Conference on Computer Vision},
  pages={23051--23061},
  year={2023}
}

@inproceedings{tumanyan2023plug,
  title={Plug-and-play diffusion features for text-driven image-to-image translation},
  author={Tumanyan, Narek and Geyer, Michal and Bagon, Shai and Dekel, Tali},
  booktitle={Proceedings of the IEEE/CVF Conference on Computer Vision and Pattern Recognition},
  pages={1921--1930},
  year={2023}
}

@article{hu2021lora,
  title={Lora: Low-rank adaptation of large language models},
  author={Hu, Edward J and Shen, Yelong and Wallis, Phillip and Allen-Zhu, Zeyuan and Li, Yuanzhi and Wang, Shean and Wang, Lu and Chen, Weizhu},
  journal={arXiv preprint arXiv:2106.09685},
  year={2021}
}

@inproceedings{chen2023diffusiondet,
  title={Diffusiondet: Diffusion model for object detection},
  author={Chen, Shoufa and Sun, Peize and Song, Yibing and Luo, Ping},
  booktitle={Proceedings of the IEEE/CVF international conference on computer vision},
  pages={19830--19843},
  year={2023}
}

@article{zhao2022word2pix,
  title={Word2pix: Word to pixel cross-attention transformer in visual grounding},
  author={Zhao, Heng and Zhou, Joey Tianyi and Ong, Yew-Soon},
  journal={IEEE Transactions on Neural Networks and Learning Systems},
  volume={35},
  number={2},
  pages={1523--1533},
  year={2022},
  publisher={IEEE}
}

@inproceedings{liu2024vgdiffzero,
  title={VGDiffZero: Text-to-image diffusion models can be zero-shot visual grounders},
  author={Liu, Xuyang and Huang, Siteng and Kang, Yachen and Chen, Honggang and Wang, Donglin},
  booktitle={ICASSP 2024-2024 IEEE International Conference on Acoustics, Speech and Signal Processing (ICASSP)},
  pages={2765--2769},
  year={2024},
  organization={IEEE}
}

\end{document}